\documentclass[a4paper]{article}

\usepackage{caption}
\usepackage{subcaption}
\usepackage{color}
\usepackage{mathtools}
\usepackage{amsmath}
\usepackage{hyperref}
\usepackage{graphicx}
\usepackage[table]{xcolor}
\usepackage{INTERSPEECH2019}

\definecolor{green}{RGB}{0,128,0}
\definecolor{blue}{RGB}{0,0,255}
\definecolor{red}{RGB}{220,20,60}
\definecolor{orange}{RGB}{255,140,0}

\title{Learning to Detect Bipolar Disorder and Borderline Personality Disorder with Language and Speech in Non-Clinical Interviews}
\name{Bo Wang$^1$$^,$$^3$, Yue Wu$^2$$^,$$^3$, Niall Taylor$^1$, Terry Lyons$^2$$^,$$^3$\\
Maria Liakata$^3$$^,$$^4$, Alejo J Nevado-Holgado$^1$, Kate E A Saunders$^1$}
\address{
  $^1$Department of Psychiatry, University of Oxford, UK\\
  $^2$Mathematical Institute, University of Oxford, UK\\
  $^3$The Alan Turing Institute, London, UK \\
  $^4$Department of Computer Science, University of Warwick, UK
  }
\email{bo.wang@psych.ox.ac.uk}

\begin{document}

\maketitle
\begin{abstract}
Bipolar disorder (BD) and borderline personality disorder (BPD) are both chronic psychiatric disorders. However, their overlapping symptoms and common comorbidity make it challenging for the clinicians to distinguish the two conditions on the basis of a clinical interview. 
% Recent studies have explored data driven approaches to automatically screen patients, incorporating features extracted from clinical interviews, showing diagnostic value for mental health conditions such as depression and bipolar disorder. 
In this work, we first present a new multi-modal dataset containing interviews involving individuals with BD or BPD being interviewed about a non-clinical topic . We investigate the automatic detection of the two conditions, and demonstrate a good linear classifier that can be learnt using a down-selected set of features from the different aspects of the interviews and a novel approach of summarising these features. Finally, we find that different sets of features  characterise BD and BPD, thus providing insights into the difference between the automatic screening of the two conditions.
\end{abstract}

\section{Introduction}
Bipolar disorder (BD) is a recurrent chronic mental health condition which occurs in approximately 1\% of the global population \cite{grande2016bipolar}. It is characterised by episodes of low and high mood which cause significant interference with everyday life. Borderline personality disorder (BPD) is characterised by a long-term pattern of constantly variable mood, self-image and behaviour. Although BD and BPD are two very different conditions they share some similar symptoms such as mood instability and impulsive behaviour \cite{ruggero2010borderline}. A recent study \cite{fornaro2016prevalence} reported the high prevalence of comorbidity between the two conditions, with up to 21.6\% of individuals with BD found to have comorbid BPD. As a result they can be difficult to distinguish, but accurate diagnosis is crucial as they require different treatment \cite{national2009borderline,national2014bipolar}. Standard diagnostic assessment involves a psychiatrist asking a series of questions about symptoms and the person has to retrospectively describe their account of these symptoms. The success of the assessment also relies on how the psychiatrist interprets both the verbal and non-verbal cues drawn from the person's responses. In this work, we aim to develop a method that extracts cues automatically from interviews conducted in a non-clinical setting, to assist the existing assessment framework, which is expensive and subjective.

Recent studies have explored data driven approaches to automatically screen patients, incorporating features extracted from multiple modalities in clinical interviews, showing diagnostic value for mental health conditions such as depression and bipolar disorder \cite{al2018detecting,aldeneh2019identifying,matton2019into,Voleti2019}. \cite{matton2019into} finds the performance of automatic mood detection to be much better in clinical interactions than in personal conversations, and there are significant differences in the features important to each type of interaction. While existing studies of BD  have focused on recognising mood episodes, the distinction between BD and BPD remains understudied. In this paper, we aim to bridge this gap by presenting a multi-modal (i.e. speech and text) dataset containing interviews in a non-clinical setting involving individuals with a diagnosis of BD or BPD, and study the automatic assessment of the two mental health conditions.

 Motivated to study the interaction between the interviewer and participant during the course of an interview from different aspects (including linguistic complexity, semantic content and dialogue structure), we investigate features extracted from different modalities. Path signatures, initially introduced in rough path theory as a branch of stochastic analysis, has been shown to be successful in a range of machine learning tasks involving modelling temporal dynamics \cite{arribas2018signature,wang2019path,kidger2019deep}. We propose to apply path signatures for summarising features extracted from each utterance, sentence and speaker-turn into interview-level feature representations, given its ability to naturally capture the order of events. By doing so, we automatically include more non-linear prior knowledge in our final feature set, which leads to effective classification, even with a simple linear classifier. 

The contributions of this work are as follows: (1) We present a new non-clinical interview dataset involving BD and BPD patients; (2), We investigate different feature types and propose using path signatures as a novel approach of summarising turn-level features; (3) We demonstrate a good linear model can be learnt for three classification tasks, and provide insights into the distinction between BD and BPD by analysing the importance of the selected features.

\section{AMoSS Interview Dataset}
% 9 participants who withdrew consent or failed to provide at least 2 months of data were excluded from further analysis.
The original Automated Monitoring of Symptoms Severity (AMoSS) study \cite{tsanas2016daily, arribas2018signature} was a longitudinal study during which a range of wearables in combination with a bespoke smartphone app were used for the daily self-monitoring of mood instability. Among the 139 participants enrolled in the study, 53 had a BD diagnosis, 33 had been diagnosed with BPD and 53 were healthy volunteers. All the diagnoses had been confirmed prior to the study using the structured clinical interview for DSM-IV (the 4th edition of Diagnostic and Statistical Manual of Mental Disorders) and the International Personality Disorder Examination (IPDE) \cite{loranger1994international}. The majority of the BD participants were euthymic while BPD participants were not in crisis but were still symptomatic as is the case with chronic experience of the condition. Exclusion criteria for BD and BPD were comorbidity of each diagnosis \cite{mcgowan2019circadian}.

62 participants were interviewed halfway through the study to gather qualitative feedback and discuss potential improvement\footnote{The study protocol was approved by the NRES Committee East of England—Norfolk (13/EE/0288), and all 62 participants consented to be interviewed and for those interviews to be recorded.}. Each participant was interviewed only once. These semi-structured, one-on-one qualitative interviews took place either in person or by telephone, conducted by 2 clinicians and 2 psychology graduates who were involved in the roll out of the AMoSS study. The natural conversations recorded between the interviewer and participant are usually within the scope of: 1) experience using the mood reporting app and the questionnaires in the app; 2) experience using different wearable devices; 3) benefits of taking part in the study and discussion of potential improvement, making the interviews semi-structured.
% \footnote{The reviewer with the most transcription experience conducts the second round reviewing to ensure the quality and consistency.}

\begin{figure}[htb]
\begin{subfigure}{.5\textwidth}
  \centering
  % include third image
  \includegraphics[width=.8\linewidth]{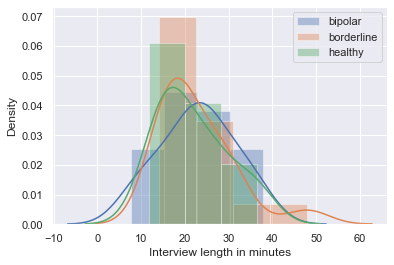}  
%   \caption{Put your sub-caption here}
  \label{fig:sub-third}
\end{subfigure}
\begin{subfigure}{.5\textwidth}
  \centering
  % include fourth image
  \includegraphics[width=.8\linewidth]{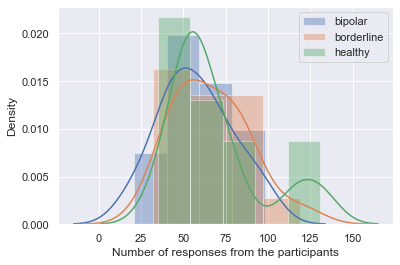}  
%   \caption{Put your sub-caption here}
  \label{fig:sub-fourth}
\end{subfigure}
\caption{Density plots of: length of the interview (\textit{top}), and number of responses from the participants (\textit{bottom})}
\label{fig:density-plots}
\end{figure}

The AMoSS interview (AMoSS-I) dataset we study here consists of 50 randomly sampled interviews that were initially transcribed by the same interviewers. The audio recordings and manually transcribed text were then aligned by a Sakoe-Chiba Band Dynamic Time Warping based forced speech alignment model in \textit{aeneas}\footnote{https://github.com/readbeyond/aeneas}. This was followed by a convolutional neural network (CNN) based noise-robust speech segmentation \cite{ddoukhanicassp2018}, generating speaker-turn-level alignments in the time domain. The manual transcripts and automatically generated time alignments were then reviewed and improved by three research assistants, where each interview was reviewed twice to ensure quality and consistency. We used finetuneas\footnote{https://github.com/ozdefir/finetuneas} as the annotation interface for reviewing. The demographic details of the participants are summarised in Table~\ref{tab:data1}. We can see both BIS-11\footnote{The Barratt Impulsiveness Scale (BIS-11) \cite{patton1995factor} is a self-report questionnaire designed to assess the personality/behavioral construct of impulsiveness. Higher BIS-11 scores are indicative of higher impulsivity.} and IPDE scores are higher among BD and BPD patients compared to controls. We also observe from the density plots in Figure~\ref{fig:density-plots} that distributions of the three user groups are very similar in interview length and number of participant responses. 
% We also do not see clear separation among the distributions of their BIS-11 scores\footnote{The Barratt Impulsiveness Scale (BIS-11) \cite{patton1995factor} is a questionnaire designed to assess the personality/behavioral construct of impulsiveness. Higher BIS-11 scores are indicative of higher impulsivity.} between BD and BPD. 
% indicating the difficulty in distinguishing one from the other.   

\begin{table}[htb]
\begin{center}
\begin{tabular}{c|c|c|c}
\hline \bf  & \bf BD & \bf BPD & \bf HC\\ \hline %BDI/BDII
%  &  & \\
% \hline
\#Interviews & 21 & 17 & 12\\
\#Room Interviews & 14 & 9 & 9\\
\#Phone Interviews & 7 & 8 & 3\\
Gender (m:f) & 7:14 & 1:16 & 3:9\\
Age (years) & $44\pm{17}$ & $34\pm{21}$ & $34\pm{18.5}$\\
BIS-11 score & $67\pm{16}$ & $76\pm{22}$ & $48.5\pm{10.25}$\\
IPDE score & $2\pm{5}$ & $16\pm{3}$ & $0\pm{0}$\\
%  &  &  & \\
\hline
\end{tabular}
\end{center}
\caption{Demographic characteristics of the three groups: Bipolar disorder (BD), Borderline personality disorder (BPD) and Healthy controls (HC). Appropriate distributions are summarised in the form of the median +/- the interquartile range.}
\label{tab:data1}
\end{table}

\begin{figure*}[h]
 \begin{center}
  \includegraphics[width=0.85\textwidth]{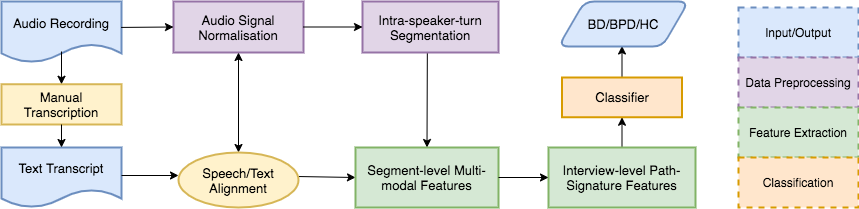}
  \caption{Data pipeline including three stages of data preprocessing (in {\color{purple}purple}), feature extraction (in {\color{green}green}) and classification (in {\color{orange}orange}).}
  \label{fig:data-pipeline}
 \end{center}
\end{figure*}

% Among the 50 interviews in AMoSS-I, 32 were recorded by placing a \todo{standard recorder?} between the interviewer and participant in the meeting room, while the remaining 18 were phone interviews in which only the interviewer's end of the phone was recorded. The duration of the interviews are ranging from 7.7 to 47.9 minutes. As summarised in Table~\ref{tab:data2}, the phone interviews are more likely to be shorter than the ones conducted in the meeting room. We also take the difference of the peak and trough values for Root-mean-square (RMS) level measured over a 10ms window, showing the difference of loudness within each audio recording. Comparing to the meeting room interviews, on average the phone interviews are shown to have significant larger difference of loudness within each one, mainly due to the way of recording. Additionally, we found the noise level in the room interviews to be higher as seen in their much lower signal-to-noise ratio of 21.16 dB, computed by Waveform Amplitude Distribution Analysis (WADA-SNR) \cite{kim2008robust}.

% The two different recording environments also resulted in different levels of clipping. As shown in Table~\ref{tab:data2}, clipping occurs much more often in the meeting room interview recordings than the phone interviews, with an average of 10.28\% of the audio signals being clipped from their maximum range. We have reduced the level of clipping by extrapolating the clipped the parts of the audio using an open-source digital audio editor \textit{Audacity}\footnote{https://www.audacityteam.org/}.

\subsection{Data Preprocessing}
% We first reduce the level of audio clipping by extrapolating the clipped the parts of the audio using an open-source digital audio editor \textit{Audacity}\footnote{https://www.audacityteam.org/}. 
Some of the interview recordings show noticeable difference of loudness between two speakers, mainly due to the mode of recording. In order to alleviate the effect of loudness difference, we scale the audio signal for each speaker turn separately, and make sure each turn is in the range of -1 and 1. We also perform intra-speaker-turn segmentation using an end-to-end voice activity detection (VAD) model \cite{Lavechin2020} trained on the DIHARD Speech Diarisation data \cite{ryant2019second}. The model extracts domain-independent features during its domain-adversarial multi-task training on DIHARD, showing better performance over the standard VAD models that do not use such domain information. This in turn allows us to extract dialogue features related to pauses in speech appearing within each speaker turn.

\section{Feature Extraction}
\label{sec:features}
We identify a set of features that are motivated by existing work in automatic mental health assessment and clinical studies of bipolar disorder symptoms. These features are selected to allow studying the interviews from different aspects
% while maintaining a level of feature compactness suited to a small dataset. They include
: lexical diversity and density, syntax, semantic content and dialogue structure. 

% \textbf{Rhythm features}:
% During these dyadic interviews, participants often exhibit changes in the rhythm of their speech when they are recollecting their experience especially of the intensive week of the study. Previous studies have used the speech rhythms features to effectively detect mood states \cite{gideon2016mood,aldeneh2019identifying}. We extract seven rhythm features from each speak-turn using algorithm proposed by Tilsen and Arvaniti \cite{tilsen2013speech} that is based on empirical mode decomposition of the vocalic energy amplitude envelope. They show these rhythm features can capture information about periodicities that likely correspond to different linguistic constructs, and thus are useful for examining rhythmicity in speech.

\textbf{Linguistic complexity features (LING)}:
Previous studies have shown language disturbances such as pressure of speech in mania and poverty of speech in depression are among the main symptoms of acute episodes in BD \cite{goodwin2007manic,weiner2019thought}. 
% while psychotic symptoms such as formal thought disorder are also common features \cite{dunayevich2000prevalence}. 
We adopt a set of linguistic complexity measures used in \cite{Voleti2019}, including measurements for lexical diversity such as \textit{moving average type-to-token ratio} (MATTR), \textit{brun\'et's index} (BI) and \textit{honor\'e's statistic} (HS); lexical density such as \textit{number of function words per word} (FUNC/W) and \textit{number of interjections per word} (UH/W); and \textit{mean length sentence} (MLS). These features are shown to be effective in distinguishing patients with schizophrenia and bipolar I disorder. We also use the dependency-based propositional idea density (DEPID), originally proposed in \cite{sirts2017idea} for measuring the rate of propositions or ideas expressed per word in spontaneous speech transcripts. 
% We denote this set of features as lexical features (\textbf{LEX}). 

In addition to the aforementioned measures we extract part-of-speech related variables including: \textit{first person pronouns} (it has been shown that people suffering from depression use more first person pronouns in written \cite{rude2004language} and spoken language \cite{zimmermann2017first}); \textit{swear words}; \textit{speech disfluencies}; \textit{filler words/phrases} such as ``okay'' and ``you know'', using the LIWC-2015 dictionary \cite{pennebaker2015development}. Finally, we add the \textit{number of absolutist words per word} (ABS/W) as a feature using a 19-word absolutist dictionary curated by clinical psychologists \cite{al2018absolute}, who found that the online anxiety, depression, and suicidal ideation forums contained more absolutist words than the control ones. Overall we extract 28 linguistic features, denoted as \textbf{LING}.
% Overall we extract 26 lexical \textbf{LEX} and syntactic \textbf{SYX} features.

%Yngve depth score, Frazier score; 

\textbf{Semantic content features (CNT)}: 
We extract content related to psychological states from the transcripts by applying relevant categories of LIWC \cite{tausczik2010psychological} such as emotions (e.g. anxiety), social processes (e.g. family) and drives (e.g. reward and risk). We also apply the empathetic concern and personal distress lexica \cite{sedoc2020learning}, and an optimism lexicon (i.e. positive future-oriented thinking), by taking the sum over all weighted words multiplying their relative frequencies in the speaker turn. Overall we use 19 content features.

\textbf{Dialogue features (DIAL)}:
We use the set of high-level turn-taking behaviour related features proposed in \cite{aldeneh2019identifying}, namely \textit{relative floor control}, \textit{turn hold offset} (i.e. short pauses that are less than half a second), \textit{number of consecutive turns} (separated by longer than half a second pauses), \textit{turn switch offset} and \textit{turn length}, per speaker turn. \textit{Turn switch offset} measures the latency between speaker turns, and it is shown that depressed people take more time to respond to questions by clinicians \cite{yu2013multimodal}. While \textit{Relative floor control} measures the percentage of time an individual controlled the conversation floor up to the time of speaking, we also add \textit{relative turn length} that measures the percentage of the length of the current turn relative to the average turn length up to the time of speaking. We compute both features in seconds as well as in number of words, following \cite{meshorer2016using}.
% \textit{Relative floor control} measures the percentage of time an individual controlled the conversation floor up to the time of speaking, which we compute in seconds as well as in number of words, following \cite{meshorer2016using}. 
% While \textit{Relative floor control} measures the percentage of time an individual controlled the conversation floor up to the time of speaking

Talking over or interrupting others, is one of the characteristics for pressure speech, which has shown to be a significant feature in bipolar mania \cite{andreasen1979thought1,andreasen1979thought2}. We use a LSTM-based overlapping speech detection model proposed in \cite{Bullock2020} to extract the number and duration of \textit{speech overlaps} in each speaker turn as features. Additionally, we add the \textit{number of words per second} per turn, as a feature representing speaking speed.

We average \textit{turn hold offset} and \textit{speech overlaps} per turn, in addition to \textit{number of consecutive turns}, \textit{relative floor control (time)}, \textit{relative floor control (words)}, \textit{relative turn length (time)}, \textit{relative turn length (words)}, \textit{turn switch offset}, \textit{turn length} and \textit{number of words per second}, which results in 11 features representing each speaker turn. 

% \textbf{Non-verbal expression features?}:
% ['sp_mean', 'n_lpauses', 'tso', 'wps', 'tl', 'lp_mean', 'ct_p', 'fcr_t', 'ovl_mean']

\subsection{Interview-level Feature Representation}
% In existing studies, the features extracted from each utterance, sentence or speaker-turn are summarised into interview-level statistics by applying a set of high-level statistical pooling functions (HSFs) such as mean and standard deviation. 
% Path signature\footnote{We refer the reader to \cite{lyons2014rough} for a rigorous introduction of path signature, and \cite{chevyrev2016primer} for a primer on its use in machine learning.}, which was initially introduced in rough path theory as a branch of stochastic analysis, has shown recent success in a range of sequence modelling tasks \cite{arribas2018signature,wang2019path,kidger2019deep}.
The theory of rough paths, developed by Lyons \cite{lyons1998differential}, can be thought of as a non-linear extension of the classical theory of controlled differential equations. The signature of a path\footnote{We refer the reader to \cite{lyons2014rough} for a rigorous introduction of path signatures, and \cite{chevyrev2016primer} for a primer on its use in machine learning.} (i.e. an ordered data stream) is a collection of $n$-fold iterated integrals such that every continuous function of the path may be approximated arbitrarily well by a linear function of its signature. Motivated by its ability to naturally capture the order of events and model temporal dynamics, we apply signature transform (SIG), which is the map from a path to its signature, to each type of the turn-level features and generate interview-level fixed-length feature representation. 
% We evaluate SIG against the popular statistical pooling functions\footnote{We apply widely used pooling functions: \textit{mean}, \textit{standard deviation}, \textit{kurtosis}, \textit{skewness}, \textit{median}, \textit{min}, \textit{max}, \textit{range}, \textit{normalised locations of min and max}, \textit{linear regression slope}, \textit{intercept} and \textit{error}.} (HSFs) in Section~\ref{results}.

\section{Experiments and Analysis}
% We use logistic regression as the classification model given our preference over interpretability and the size of our dataset.
% Following previous works we choose to evaluate our model using leave-one-participant-out cross validation (LOOCV). 
Following previous work we chose leave-one-participant-out (LOOCV) as the evaluation scheme, and logistic regression as the classification model given our preference over interpretability and the size of our data. For each fold, we first apply signature transform to each type of turn-level features, and keep only the first three levels of the path signature\footnote{We use iisignature Python library, \url{https://pypi.org/project/iisignature/}, and set the maximal order to which iterated integration is performed in signature to be 3.}. To avoid overfitting, we conduct feature selection on signature-transformed interview-level features through computing Pearson Correlation Coefficients (PCC) with the IPDE scores on the training data and retain the features with $p$-values less than 0.001. This results in a small number of features. The selected features are then fed to the classifier for 3 separate binary tasks: (1), BD vs. healthy controls, (2), BPD vs. healthy controls, and (3), BD vs. BPD patients. We conduct three separate experiments, extracting features from the speech of each participant and interviewer respectively, as well as the whole interview (as a sequence of turns) without speaker identification (denoted as `Both')\footnote{For the `Interviewer' and `Both' experiments, we increase the p value threshold to 0.002.}.

% Then I retain the interview-level (signature-transformed) features that have statistically significant Pearson Correlation Coefficients (PCC) with the IPDE scores on the training data. Thus those features that have $p$-values less than 0.001, are removed\footnote{This results in a small number of features, usually in single digit, to avoid overfitting.}.

\subsection{Analysis of the selected features}
Five highly ranked and most commonly selected features from each task are briefly summarised in Table~\ref{tab:feat-ranking} as examples. Each interview-level feature is represented as a linear combination of the original turn-level features. We see over half of the selected features are volume integrals, i.e. they are triple integrals of three turn-level features, while the rest are double integrals which give the \textit{Le\'vy area}. We notice most all of the representative features are from the linguistic category (many are part-of-speech tags), especially for \textit{H vs. BD} and \textit{H vs. BPD}, showing the significance of the structure in the interviews. This is inline with the finding in \cite{matton2019into}, and extends to nonclinical interviews conducted partially by students. 

We also notice the appearances of nonfluencies (Nonflu.) especially in combination with conjunctions (CONJ) in the detection of \textit{H vs. BPD}. The use of the absolute words (ABS) in combination with common adverbs (ADV) and negations (NEG) or article words (e.g. a, an, the) are selected for the two classification tasks involving BPD. Two interview-level features \textit{(DEPID, MATTR, BI)} and \textit{(BI, MATTR, MLS)} are shown to be the most commonly selected for the two tasks involving BD.

\begin{table}[htb]
\begin{center}
\scalebox{0.85}{
\begin{tabular}{c|c|c}
\bf H vs BD & \bf H vs BPD & \bf BD vs BPD \\ \hline
\color{blue}\textit{(DEPID, MATTR, BI)} & \color{blue}\textit{(Nonflu., CONJ)} & \color{blue}\textit{(BI, MATTR, MLS)} \\
\color{blue}\textit{(Nonflu., Verbs)} & \color{blue}\textit{(ABS, ADV, Articles)} & \color{blue}\textit{(We, PREP)} \\
\color{blue}\textit{(PPRO, CONJ, CONJ)} & \color{green}\textit{(WPS, SP\_avg, RFC\_t)} & \color{blue}\textit{(PREP, We)} \\
\color{blue}\textit{(NEG, AUXV, NEG)} & \color{blue}\textit{(CONJ, Nonflu.)} & \color{blue}\textit{(ABS, ADV, NEG)} \\
\color{blue}\textit{(PPRO, Swear, Verbs)} & \color{blue}\textit{(You, Verbs, Nonflu.)} & \color{red}\textit{(SOC, DRI, DRI)} \\
\hline
\end{tabular}}
\end{center}
\caption[dummy caption]{Top-5 highly ranked and most commonly selected features during LOOCV. Feature selection is based on correlation between each feature and the IPDE scores of the training samples. Features belong to the linguistic category are colored in {\color{blue}blue}; dialogue feautres are in {\color{green}green} and content features are in {\color{red}red}.\footnotemark}
\label{tab:feat-ranking}
\end{table}
\footnotetext{PPRO: personal pronouns; PREP: prepositions; ADV: adverbs; AUXV: auxiliary verbs; CONJ: conjunctions; NEG: negations; Nonflu.: nonfluencies; We: first-person plural; Swear: swear words; MLS: mean length sentence; WPS: number of words per second; SP\_avg: average length of short pauses; RFC\_t: relative floor control (time); SOC: social processes; DRI: drives.}

\subsection{Results and Discussion}
\label{results}
The results for the classification tasks are summarised in Table~\ref{tab:results}. Using the (late) fusion of linguistic, dialogue and content features with signature transform, we obtain a AUROC of 0.810 in H/BD, 0.733 in H/BPD and 0.817 in BD/BPD. We notice the result in H/BPD is significantly worse than the other two tasks. We leave its investigation to future work, we think the fewer data samples and varying recording quality may have contributed to the worse performance. When we model from the speaking segments of the interviewers, we obtain very poor performance. As the purpose of the interviews were merely to understand the individual's experience of taking part in the AMoSS study rather than establishing their mental state at the time of interview, it is no surprise that features extracted from the interviewers have very weak discriminative power. We also believe having different interviewers impacted negatively on the classifications. Modelling the interviews as a sequence of utterances also resulted in much worse performance than learning from the participants alone. 

\begin{table}[htb]
\centering
\scalebox{0.9}{
\begin{tabular}{c|ccc}
\hline
\multicolumn{1}{c}{} & \multicolumn{3}{c}{\bf AUROC} \\
\bf Subject & \bf H/BD & \bf H/BPD & \bf BD/BPD\\ \hline
Participant & \textbf{0.810} & \textbf{0.733} & \textbf{0.817}\\
% Participant & SIG & Global & 0.560? & 0.564? & 0.574?\\
% Participant & HSF & Person & 0.452 & 0.674 & 0.604\\
% Participant & HSF & Global & 0.452 & 0.615 & 0.604\\

Interviewer & 0.304 & 0.473 & 0.231\\
% Interviewer & SIG & Global & 0.506 & 0.431 & 0.583\\
% Interviewer & HSF & Person & 0.417 & 0.556 & 0.622\\
% Interviewer & HSF & Global & 0.369 & 0.444 & 0.651\\

Both & 0.494 & 0.431 & 0.657\\
% Both & SIG & Global & 0.571 & 0.439 & 0.675\\
% Both & HSF & Person &  &  & \\
% Both & HSF & Global & 0.435 & 0.686 & 0.639\\

%  &  &  &  &  & \\
%  &  &  &  &  & \\
\hline 
\end{tabular}}
\caption{Classification results for three binary tasks: H vs. BD, H vs. BPD and BD vs. BDP, using logistic regression. Results shown are average AUROCs across all interviews.}
\label{tab:results}
\end{table}

We also conduct ablation experiments to examine how performance changes after removing each feature type. As seen in Table~\ref{tab:feat-ablation}, the linguistic features (LING) are the biggest contribution in all three tasks. As a consequence, we have to increase the p-value feature selection threshold from 0.001 to 0.005 to have any feature for classification, if we remove LING. A sharp performance drop is then observed removing LING in in all three tasks. If we remove both LING and dialogue features (DIAL), the results get even worse. If we exclude both DIAL and content features (CNT), we still get reasonably good performance without significant drop in AUROC.
% As a consequence, no feature is statistical significant enough to be selected by the model if we remove LING (represented by the grey cells in the table). We also see a sharp performance drop by removing LING in H vs. BD. As for BD vs. BPD, the exclusion of the linguistic features resulted in AUROC dropping from 0.811 to 0.569. Excluding both the content (CNT) and linguistic features also resulted the drop of AUROC to 0.646. 
It's also worth noticing the ineffectiveness of the content features (CNT) apart from in BD vs. BPD, possibly due to the semi-structured nature of the interviews and the questions asked fall within the same scope. Given that the majority of BD and BPD participants were clinically stable this may also account for the relatively poor distinction between the groups using CNT features that are related to the psychological states from responses by the participants. 
% Given that the majority of BD and BPD participants were clinically stable this may account for the relatively poor distinction between the groups but the different recording environments and interviewers may also have impacted the structure of the dialogues in some interviews. 

\begin{table}[htb]
\begin{center}
\scalebox{0.9}{
\begin{tabular}{lccc}
\hline \bf Features & \bf H vs BD & \bf H vs BPD & \bf BD vs BPD\\ \hline
All & 0.810** & 0.733** & 0.817**\\
All-CNT & 0.810** & 0.733** & 0.787**\\
All-DIAL & 0.768** & 0.733** & 0.811**\\
All-LING & 0.625* & 0.578* & 0.669*\\
% All-LEX & 0.804 & 0.733 & 0.651\\
% All-SYX & 0.738 & \cellcolor{gray} & 0.669\\
All-LING-CNT & 0.642* & 0.703* & 0.604*\\
All-LING-DIAL & 0.442* & 0.429* & 0.550*\\
% All-LEX-DIAL & 0.762 & 0.733 & 0.651\\
% All-LEX-CNT & 0.804 & 0.733 & 0.734\\
% All-DIAL-SYX & 0.762 & \cellcolor{gray} & 0.669\\
% All-LEX-SYX & \cellcolor{gray} & \cellcolor{gray} & 0.569\\
All-CNT-DIAL & 0.768** & 0.733** & 0.763**\\
% All-CNT-SYX & 0.758 & \cellcolor{gray} & 0.646\\
% All-CNT-DIAL-SYX & 0.758 & \cellcolor{gray} & 0.646\\
% All-CNT-DIAL-LEX & 0.762 & 0.733 & 0.734\\
%  &  &  & \\
\hline
\end{tabular}}
\end{center}
\caption{Feature ablation results (AUROC) for each task, e.g. the final row in the table shows the result for using linguistic features (LING) only. p-value used for feature selection: `**'$<$0.001; `*'$<$0.005.}
\label{tab:feat-ablation}
\end{table}
\vspace{-4mm}
% \begin{table}[htb]
% \begin{center}
% \scalebox{0.90}{
% \begin{tabular}{lccc}
% \hline \bf Features & \bf H vs BD & \bf H vs BPD & \bf BD vs BPD\\ \hline
% All & 0.798** & 0.887** & 0.710**\\
% All-CNT & 0.798** & 0.716** & 0.710**\\
% All-DIAL & 0.690** & 0.858** & 0.739**\\
% All-LING & 0.571* & 0.645* & 0.503*\\
% All-LING-CNT & 0.506* & 0.745* & 0.580*\\
% All-LING-DIAL & 0.315* & 0.490* & 0.710*\\
% All-CNT-DIAL & 0.690** & 0.640** & 0.739**\\
% \hline
% \end{tabular}}
% \end{center}
% \caption{Feature ablation results (AUROC) for each task, e.g. the final row in the table shows the result for using linguistic features (LING) only. p-value used for feature selection: `**'$<$0.001; `*'$<$0.005.}
% \label{tab:feat-ablation}
% \end{table}

% \begin{table}[htb]
% \begin{center}
% \begin{tabular}{cc|cc|cc}
% \multicolumn{2}{c}{\bf H vs BD} & \multicolumn{2}{c}{\bf H vs BPD} & \multicolumn{2}{c}{\bf BD vs BPD}\\
% \bf Feature & \bf Rank & \bf Feature & \bf Rank & \bf Feature & \bf Rank\\ \hline
% a & b & c & d & e & f\\
%  &  &  &  &  &  \\
%  &  &  &  &  &  \\
%  &  &  &  &  &  \\
% \hline
% \end{tabular}
% \end{center}
% \caption{Feature rankings}
% \label{tab:feat-ranking}
% \end{table}

\section{Conclusions}
In this paper, we demonstrate the potential of using features extracted from language and speech in non-clinical interviews to assist the assessment of bipolar disorder BD and borderline personality disorder BPD, which is challenging for clinicians to distinguish. 
% It is crucial to diagnose the two conditions accurately so the patients can have appropriate treatment. While many machine learning based studies learn from clinical interviews to automatically screen mental health conditions, the detection of BD and BPD is still understudied.
We first presented a non-clinical interview dataset, named AMoSS-I, conducted partially by psychology graduates, for the task of detecting BD and BPD. We demonstrated good performance in three classification tasks using down-selected features and a new way of summarising these features based on path signatures. Lastly, we showed the importance of linguistic features in all three tasks and the benefits of feature fusion from different modalities. For future work, we plan to learn acoustic features, and investigate the effect of acoustic properties of the interviews and the impact of recording environments.

\section{Acknowledgements}
This work was supported by the MRC Mental Health Data Pathfinder award to the University of Oxford [MC\_PC\_17215], by the NIHR Oxford Health Biomedical Research Centre and by the The Alan Turing Institute under the EPSRC grant EP/N510129/1. We would also like to thank Zakaria Aldeneh and Kairit Sirts for sharing code and thoughts, and to Priyanka Panchal and Rota Silva for conducting the interviews. The views expressed are those of the authors and not necessarily those of the NHS, NIHR or the Department of Health.

\bibliographystyle{IEEEtran}

\bibliography{mybib}

\end{document}